\documentclass[twoside]{article}
\usepackage[preprint]{aistats2027}

\usepackage{amsmath,amssymb,amsthm}
\usepackage{hyperref}
\usepackage{dsfont}
\usepackage{enumitem}
\setlist{itemsep=1pt, parsep=0pt, topsep=3pt}
\usepackage{graphicx}   
\usepackage[style=numeric, isbn=false, url=false, eprint=false]{biblatex}
\usepackage{booktabs}

\defbibheading{aistats}{\subsubsection*{References}}   
\graphicspath{{experiments/}}

\newtheorem{theorem}{Theorem}
\newtheorem{lemma}[theorem]{Lemma}
\newtheorem{corollary}[theorem]{Corollary}
\newtheorem{proposition}[theorem]{Proposition}
\theoremstyle{definition}

\theoremstyle{remark}

\DeclareMathOperator*{\argmin}{arg\,min}

\begin{document}

\twocolumn[
\aistatstitle{D-SLR: The Disjoint Row-Sparse plus Low-Rank Decomposition}
\aistatsauthor{ Vincent Szolnoky}
\aistatsaddress{ Department of Mathematical Sciences \\
  Chalmers University of Technology
  }]

\begin{abstract}
  Compressing a matrix for reconstruction still defaults to the truncated SVD, approximating the data with a single low-rank structure. It is common to reduce the residual further by adding an overlapping row-sparse component, but methods that solve this joint problem often require iterative solvers and tuning of regularization parameters. We propose the Disjoint Row-Sparse plus Low-Rank (D-SLR) decomposition, a closed-form drop-in for the truncated SVD that improves or exactly matches it. D-SLR restricts rows to either being stored verbatim or approximated by the low-rank fit, never both. Under squared error this restriction costs nothing: the joint optimum is attainable disjointly with fewer parameters at every non-trivial rank and stored row count (shape). With zero stored rows D-SLR reduces to the truncated SVD, so it never does worse at equal cost. The algorithm scores the entire error-versus-parameters tradeoff, and the solution is chosen afterwards by a supplied error target or parameter count, or by a selection rule. The grid and solution together cost three SVDs, with no tuning or regularization. We derive an assumption-free, a-posteriori lower bound on the error at every shape, giving each solution a computable certificate on the potential gain of any other choice of rank and stored rows. Experiments on synthetic and real data (LLM embedding tables, network traffic, hyperspectral images) confirm the gains and quantify the certificate.

\end{abstract}


\section{Introduction}
The truncated SVD is the default way to compress a matrix $W$ when nothing else is known about it. It fits every row with one shared low-rank basis, which works well when the rows agree on that basis. When a group of rows does not, it pulls the shared basis away from the majority, and the rank must grow to keep the error down. Since the misfit lives in whole rows, a common fix is to add a row-sparse component $S$ that absorbs them, leaving the low-rank component $L$ to capture the rest. This gives the problem
\begin{equation}\label{eq:slrobj}
  \begin{aligned}
    \min_{L, S}\ & \|W - L - S\|^2_F \\
    \text{s.t.}\ & \operatorname{rank}(L) \leq r, \\
    & S \text{ has at most $k$ non-zero rows.}
  \end{aligned}
\end{equation}
Problem \eqref{eq:slrobj} is hard for two reasons. It is non-convex and combinatorial, because the rank bound and the row count are both cardinality constraints. And $L$ and $S$ may overlap, so a row can be split between them, which couples the two components. A possible remedy is the convex relaxation
\begin{equation}\label{eq:convrelax}
  \min_{L, S}\  \|W - L - S\|^2_F + \lambda_1 \|L\|_\ast + \lambda_2 \|S\|_{2, 1},
\end{equation}
where $\|\cdot\|_\ast$ is the nuclear norm and $\|\cdot\|_{2,1}$ sums the row norms. It is convex, but its penalties are non-smooth and the components stay coupled, so it is solved iteratively, and $\lambda_1$ and $\lambda_2$ must be tuned.

D-SLR adds one requirement. Each row is either stored verbatim in $S$ or approximated by $L$, \textbf{never both}. This removes the coupling, but it looks like a concession, because it shrinks the feasible set of \eqref{eq:slrobj}. With row sparsity and squared error it is not, and removing whole rows leaves a plain SVD to solve. This is why we use rows rather than entries. We make three contributions.

\begin{enumerate}

\item \textbf{Disjointness is free.} At every rank and stored-row count, the optimum of \eqref{eq:slrobj} is attainable by a disjoint solution, which needs fewer parameters than the overlapping one, whenever both rows and rank are used (Proposition~\ref{prop:jointequiv}). With no stored rows D-SLR is the truncated SVD, so at equal parameter count it is never worse than the SVD. This makes D-SLR a drop-in for the truncated SVD.

\item \textbf{An assumption-free certificate.} From $W$ alone we lower-bound the error of every row-sparse solution (Corollary~\ref{coro:univfloor}). A block-wise construction then tightens this floor, never weakening it, by applying it to blocks of rows, which in our tests shrank the certified gap by an order of magnitude (Theorem~\ref{thm:splitfloor}). Adding a price per parameter stored turns this into a bound on how much any rival row-sparse solution could improve the score of the returned one. A bound of zero proves the returned solution is a global minimiser among row-sparse solutions, and a positive bound says how much room is left and at which ranks and row counts. Unlike a-priori guarantees, it is computed after the fit for the given matrix and assumes no model of the data.

\item \textbf{A closed-form algorithm with no tuning.} Because rows are disjoint, a solution is just a set of stored rows and an SVD of the rest. This lets us attack \eqref{eq:slrobj} directly, scoring every rank and row count with a closed-form upper bound on the error, at the cost of three SVDs and no iterations (Section~\ref{sec:algo}). Together with the certificate's floor, these scores bracket the best achievable error at every parameter count. A solution is then picked by an error target, a parameter count or a selection rule such as EBIC, so neither the rank nor row count has to be chosen in advance. Because the scores use an upper bound, the final pick can optionally refit the top few solutions on the exact error and keep the best, at one SVD per refit.

\end{enumerate}

Together, these make D-SLR a drop-in for the truncated SVD that is never worse at equal cost, costs three SVDs, and states how far it could be from the best.

On synthetic matrices built to stress each part of the method, D-SLR matched or beat iterative methods at a small fraction of their cost, and in most cases the certificate proved its score was within a few percent of the best achievable. On LLM embedding tables, when retaining only half the table's parameters, D-SLR changed about ten times fewer of the model's predictions than the truncated SVD, and matched or improved on storing the heaviest rows with the best row count chosen on the exact error.

\section{Related work}

\paragraph{Sparse-plus-low-rank models.}
Sparse-plus-low-rank decompositions are best known from robust PCA. Principal component pursuit \cite{candes2011robust} splits a matrix into a low-rank component and an overlapping, entrywise sparse component. Later work added dense noise \cite{zhou2010noisy} and replaced the convex program with faster non-convex projections \cite{netrapalli2014rpcaproj}. Closest to D-SLR in form is Outlier Pursuit (OP) \cite{xu2012op}, whose sparse component is row-sparse. These methods assume the matrix was generated as a low-rank component plus sparse perturbations, and their goal is to recover either of the components or both. D-SLR has a different goal. It approximates the given matrix as closely as possible for its parameter count, which a recovery method does not optimise. A few methods do target the approximation error, either directly at a fixed rank and sparsity \cite{zhou2011godec} or weighted by input activations to compress LLM weights \cite{zhang2025oats}. Both keep an overlapping, entrywise sparse component and alternate between the two components. LoSparse \cite{li2023losparse} instead trains a low-rank component plus a column-sparse component on the task loss. All of these methods are iterative, and each run returns one solution for one setting of its penalty weights, or of its rank and sparsity. Tracing the trade-off between error and parameter count means solving again for every setting. D-SLR's search is closed form, so a single pass scores every rank and row count at once, for three SVDs.

\paragraph{Low-rank approximations for compression.}
The truncated SVD gives the best low-rank approximation of any matrix in Frobenius norm \cite{eckart1936svd}. It is fast, closed form and assumes nothing about the data, which has made it a default compressor across many fields. In imaging it has been used from early image storage to MRI and hyperspectral data \cite{andrews1976image, mcgivney2014mri, zhang2012hyper}. In the sciences it condenses seismic, geothermal and single-cell data \cite{brankovic2021seismic, white2024geothermal, nicol2025rna} and underlies reduced-order modelling \cite{costanzo2025rom}. In machine learning it compresses neural network weights \cite{denton2014nn}, most recently in large language models \cite{saha2024caldera, wang2025svdllm}. D-SLR is also closed form and assumption-free, at the cost of a few SVDs, so it can serve as a drop-in for the SVD in any of these settings.

\paragraph{Optimality certificates.}
Guarantees in low-rank modelling are usually a priori. Dual certificates for convex relaxations prove recovery under an assumed model of the data \cite{candes2009complete, gross2011sparse}, and analyses of non-convex solvers bound the error over a model class \cite{netrapalli2014rpcaproj}. Both are properties of the method, fixed before the matrix is seen. The certificate of Theorem~\ref{thm:universal} is instead computed from $W$ after the fit and assumes no structure in $W$, like the gap reported by a branch-and-bound solver. \textcite{bertsimas2023slr} obtain such gaps for entrywise sparse plus low rank by branch-and-bound, for one rank and sparsity at a time and at a cost that grows quickly with the matrix size. Our certificate is closed form. It lower-bounds the error of every row-sparse solution at every rank and row count at once, and when positive it says where a better solution could still exist.

\paragraph{Row selection and model selection.}
Choosing a subset of rows is also central to row sampling and coresets \cite{drineas2008cur, jubran2019coresets}. There the chosen rows stand in for the whole matrix, and the rest are approximated through them. D-SLR does the opposite. It stores the rows that fit the shared basis worst and keeps them out of that basis. Choosing the rank and row count is a model-selection problem. We adapt the extended BIC \cite{chen2008ebic, luo2013ebic}, which is built for very many candidate models, as in our grid of ranks and row counts. Since the score uses a single price per parameter, other penalties such as MDL \cite{rissanen1978mdl} or data-driven penalties \cite{birge2007selection} can be used in its place.

Table~\ref{tab:positioning} summarises how these methods compare to D-SLR.


\begin{table*}[t]
  \centering
  \footnotesize
  \setlength{\tabcolsep}{3pt}
  \caption{D-SLR against methods of similar form. D-SLR is the only method with a sparse component that, like
    the truncated SVD, needs no tuning or iteration and gives the full path in one solve. {\footnotesize Recovery methods recover an assumed underlying component, and reconstruction methods
    approximate the given matrix. \textbf{Sparse}: structure of the sparse component, none if there is no sparse
    component. \textbf{Overlap}: whether the sparse and low-rank components may share entries. \textbf{Tuned}: a
    penalty weight or threshold must be set, beyond choosing the size (default: a universal value is given). \textbf{Iter.}: solved iteratively.
    \textbf{Full path}: one solve gives the solution at every rank and sparsity, so a selection rule needs
    no re-solving. \textbf{Guarantee}: type of guarantee on the returned solution. \textbf{Model}: whether
    the guarantee assumes a model of the data.}}
  \label{tab:positioning}
  \smallskip
  \begin{tabular}{@{}llllllll@{}}
    \toprule
    Method & Sparse & Overlap & Tuned & Iter. & Full path & Guarantee & Model \\
    \midrule
    \multicolumn{8}{@{}l}{\emph{Recovery}} \\
    PCP \cite{candes2011robust} & entries & yes & default & yes & no & recovery & yes \\
    Nonconvex RPCA \cite{netrapalli2014rpcaproj} & entries & yes & default & yes & no & recovery & yes \\
    Outlier Pursuit \cite{xu2012op} & rows & yes & yes & yes & no & recovery & yes \\
    RSR \cite{lerman2015reaper} & none & -- & no & yes & no & recovery & yes \\
    Trimmed PCA \cite{mateos2012rpca} & rows & yes & yes & yes & no & stationary point & no \\
    \midrule
    \multicolumn{8}{@{}l}{\emph{Reconstruction}} \\
    Truncated SVD \cite{eckart1936svd} & none & -- & no & no & yes & global optimum & no \\
    Coresets \cite{drineas2008cur, jubran2019coresets} & none & -- & no & no & no & error bound &
                                                                                                          no \\
    GoDec \cite{zhou2011godec} & entries & yes & no & yes & no & local convergence & no \\
    Discrete SLR \cite{bertsimas2023slr} & entries & yes & yes & yes & no & certified gap (small $n$) & no
    \\
    Convex relax.\ \eqref{eq:convrelax} & rows & yes & yes & yes & no & relaxed optimum & no \\
    \textbf{D-SLR} & \textbf{rows} & \textbf{no} & \textbf{no} & \textbf{no} & \textbf{yes} & \textbf{score
                                                                                              certificate} & \textbf{no} \\
    \bottomrule
  \end{tabular}
\end{table*}

\section{Problem formulation}\label{sec:problem}

\paragraph{Notation.}
Let \( W \in \mathbb{R}^{n \times m} \) be the input matrix and $\mathcal{N}$ its set of rows. For any matrix $X$ and set of rows $K$, $X_K$ denotes the rows of $X$ indexed by $K$. Write \( \sigma_1(X) \ge \sigma_2(X) \ge \dots \) for the singular values of $X$, with \( \sigma_i(X) = 0 \) beyond the smaller dimension of $X$. Define
\begin{align}
  T(X; r) &= \sum_{i > r} \sigma_i(X)^2, \label{eq:tail} \\
  R(X; k) &= \text{sum of the $k$ largest squared} \notag \\
  &\phantom{=}\ \text{row norms of $X$}. \label{eq:rownorms}
\end{align}
By the Eckart--Young theorem, \( T(X; r) = \|X - X_r\|_F^2 \), where $X_r$ is the best rank-$r$ approximation of $X$ \cite{eckart1936svd}. So $T$ is the energy a rank-$r$ fit leaves behind. When the matrix argument is omitted, as in \( T(r) \), it is the matrix under discussion, usually $W$.

\paragraph{Descriptions and solutions.}
A \emph{description} is a pair \( (r, K) \) of a rank \( 0 \le r \le \min\{n, m\} \) and a set \( K \subseteq \mathcal{N} \) of rows to store verbatim. Its \emph{shape} is \( (r, k) \) with \( k = |K| \), and \( B = \mathcal{N} \setminus K \) is the set of unstored rows. The description defines the D-SLR \emph{solution} \( \hat{W} = S + L \), where $S$ equals $W$ on the rows $K$ and is zero elsewhere, and $L$ is the best rank-$r$ approximation of $W_B$ on the rows $B$ and is zero on $K$. Its \emph{error} is
\begin{equation}\label{eq:reconerr}
  E(r, K) = \|W - S - L\|_F^2 = T(W_B; r),
\end{equation}
and for any matrix $X$ we write \( E(X; r, K) \) for the error of \( (r, K) \) on $X$.
A description says which rows to store and which rank to use, and the solution is the matrix it produces. The shape only counts the two. Many descriptions share a shape, and their cost depends on the shape alone. Where the pairs pile up, we write \( y = (r, K) \) for a description and \( s = (r, k) \) for its shape, and evaluate functions of descriptions, such as $E$, at $y$ and functions of shapes, such as $C$, at $s$.

\paragraph{Cost.}
Storing $W_K$ takes \( km \) parameters, and storing the rank-$r$ approximation of the \( n - k \) unstored rows takes \( r(n - k + m) \). A shape therefore costs
\begin{equation}\label{eq:cost}
  C(r, k) = km + r(n - k + m),
\end{equation}
with the rank term dropped when \( k = n \) and nothing is left to approximate. The cost counts what the shape reserves, whatever values are stored. We do not count the indices of the stored rows. They are $k$ integers next to the \( km \) stored values, so including them would raise the cost by a factor of at most \( 1 + 1/m \).

Before looking at any data, the cost alone already tells us where D-SLR should help most. One more stored row adds \( m - r \) parameters, while one more rank adds \( n - k + m \). A row is therefore always cheaper than a rank, because storing it also removes it from the low-rank factor. For small $r$ and $k$ a rank costs about \( 1 + n/m \) rows, so the taller the matrix (\( n \gg m \)), the more rows the same parameters can buy. D-SLR therefore tends to work best on tall matrices, although whether a row actually lowers the error more than a rank still depends on the data.

\paragraph{Problem.}
Every description gives a point of cost and error. The best trade-off between the two is the \emph{frontier}
\begin{equation}\label{eq:frontier}
  F(c) = \min \bigl\{ E(r, K) : C(r, |K|) \le c \bigr\},
\end{equation}
the lowest error reachable within $c$ parameters. Picking a solution under a parameter limit means reading $F$ at that limit, and picking one for an error target means finding the smallest $c$ that reaches it. To pick a solution without either, we price each parameter at \( \lambda > 0 \) in squared-error units and minimise the \emph{score}
\begin{equation}\label{eq:score}
  Q(r, K) = E(r, K) + \lambda C(r, |K|).
\end{equation}
The minimiser is \( (\hat{r}, \hat{K}) = \argmin_{r, K} Q(r, K) \), and Section~\ref{sec:algo} sets $\lambda$ with EBIC.

Given a returned solution, could another rival solution have a lower error at the same cost, or a lower score? If so, then \emph{potentially} by how much, and where in the solution space could it lie? And how can one efficiently compute the frontier and the score in the first place? The rest of the paper answers these in turn. Section~\ref{sec:lossless} first shows that nothing is lost by forbidding overlap. Section~\ref{sec:universal} lower-bounds the error at every shape, which gives a floor under the frontier and a certificate for the score, and Section~\ref{sec:blockwise} tightens it with blocks of rows. Section~\ref{sec:algo} gives the algorithm and Section~\ref{sec:experiments} the experiments. Proofs and further results are in the appendix.

\section{Disjointness is lossless and cheaper}\label{sec:lossless}
Forbidding overlap shrinks the feasible set of \eqref{eq:slrobj}, so does D-SLR give anything up? We now show that it does not. Take any solution of \eqref{eq:slrobj} where $S$ and $L$ overlap, and call the rows where $S$ is non-zero $K$. On those rows, $S$ can at best absorb whatever $L$ leaves behind. But why absorb only the leftover? $S$ can just as well store the entire rows $W_K$, and $L$ can drop them altogether. The error on $K$ then becomes zero, $L$ is free to fit the remaining rows $W_B$ as well as it can, and the rows of $L$ inside $K$ no longer need to be stored. This is exactly the disjoint solution of \( (r, K) \).

\begin{proposition}[Disjointness is lossless and cheaper]\label{prop:jointequiv}
  Fix $r$ and $k$. Let \( L \in \mathbb{R}^{n \times m} \) have \( \operatorname{rank}(L) \le r \), and
  let \( S \in \mathbb{R}^{n \times m} \) have all its non-zero rows inside a set $K$ with \( |K| = k
  \). Storing $L$ and $S$ separately costs \( r(n + m) + km \) parameters. Then the disjoint solution of
  \( (r, K) \) is no worse and no more expensive,
  \begin{align}
    \|W - L - S\|^2_F &\ge E(r, K), \label{eq:lossless} \\
    r(n + m) + km &\ge C(r, k), \label{eq:cheaper}
  \end{align}
  and \eqref{eq:cheaper} is strict whenever \( r > 0 \) and \( k > 0 \). In particular, the optimum of
  \eqref{eq:slrobj} equals \( \min_{|K| = k} E(r, K) \), so it is attainable by a disjoint solution.
\end{proposition}

So the overlap that couples $L$ and $S$ is not required to reach the optimum. An overlapping solution can match the disjoint one in error, but it spends extra parameters on rows that $S$ can cover anyway. The lossless part \eqref{eq:lossless} is already known in a slightly different form from the RPCA literature, where it shows that the minimising subspace is the same with and without overlap \cite{mateos2012rpca}. Here we state it for every shape and price both solutions with \eqref{eq:cheaper}. No restriction is placed on how $L$ and $S$ are built from $W$, so the result covers every overlapping solution with a row-sparse $S$. It does not extend to an entrywise sparse $S$, because zeroing single entries of $L$ generally raises its rank, so $L$ cannot simply step aside where $S$ stores values.

\section{A universal certificate valid for any matrix}\label{sec:universal}
For a subspace \( V \subseteq \mathbb{R}^m \) with \( \dim V = r \) and orthogonal projector \( P_V \), the error of the best rank-$r$ approximation of a matrix $X$ can be written as
\begin{equation}\label{eq:subspacemin}
  \min_{\dim V = r} \Bigl[ \|X\|^2_F - \|X P_V\|^2_F \Bigr] = T(X; r).
\end{equation}
With \( X = W_B \), the error \( E(r, K) \) is the smallest total squared distance from the unstored rows to any $r$-dimensional subspace. It depends on two things, which rows are stored and which subspace is used. We derive two lower bounds by attacking \eqref{eq:subspacemin} from either side. Each one forgets one of the two things and assumes the worst case in its place.

First we forget which rows are stored. Whichever they are, \( \hat{W} = S + L \) is a matrix of rank at most \( r + k \), because $S$ has $k$ non-zero rows and $L$ has rank at most $r$. So $\hat{W}$ can never be closer to $W$ than the best rank-\( (r + k) \) approximation.
\begin{lemma}[Spectral floor]\label{lem:specfloor}
  For all $r$, $k$ and every $K$ with \( |K| = k \), \( E(r, K) \ge T(r + k) \).
\end{lemma}

Next we do the opposite and forget which subspace is used. Storing a row removes its error, but that error is never more than the row's own energy. So compared with storing no rows at all, storing $k$ rows can lower the error by at most the energy of the $k$ heaviest rows.
\begin{lemma}[Row-energy floor]\label{lem:rowfloor}
  For all $r$, $k$ and every $K$ with \( |K| = k \), \( E(r, K) \ge T(r) - R(k) \).
\end{lemma}

Lemma~\ref{lem:specfloor} follows from Eckart--Young applied at rank \( r + k \), and Lemma~\ref{lem:rowfloor} from bounding what each stored row can remove from the error. Both floors hold at once, so we keep whichever is larger. This gives a floor that holds for every description at every shape, which is what the certificate below is built on.
\begin{corollary}[Universal floor]\label{coro:univfloor}
  For any matrix $X$ and all $r$, $k$, define
  \[
    \underline{E}(X; r, k) = \max \bigl\{ T(X; r + k),\ T(X; r) - R(X; k),\ 0 \bigr\}.
  \]
  Then
  \begin{enumerate}
    \item \( E(X; r, K) \ge \underline{E}(X; r, k) \) for every $K$ with \( |K| = k \), and
    \item \( \underline{E}(X; r, k) \ge T(X; r + k - j) - R(X; j) \) for every \( j \in \{0, \ldots, k\} \).
  \end{enumerate}
\end{corollary}
The second item answers a question one might have. Could a mix of the two be better, treating some of the stored rows as extra ranks and the rest as heavy rows? Each such mix is also a valid floor, but none of them beats the larger endpoints, so nothing is lost by keeping only the two.

Because the floor holds at every shape, it also bounds the frontier \eqref{eq:frontier} from below,
\begin{equation}\label{eq:frontierfloor}
  F(c) \ge \min \bigl\{ \underline{E}(r, k) : C(r, k) \le c \bigr\}.
\end{equation}
This holds with any valid floor in place of \( \underline{E} \), such as the split floor of Section~\ref{sec:blockwise}, which gives the lower bounds in our experiments.

The floor also lets us certify a single returned description \( \hat{y} \) with shape \( \hat{s} \) and score \( \hat{q} = Q(\hat{y}) \). We say a rival \( y' \) with shape \( s' \) \emph{beats} it if \( Q(y') < \hat{q} \). Writing out the score, this happens exactly when \( E(y') < E(\hat{y}) + \lambda ( C(\hat{s}) - C(s') ) \). So a cheaper rival may be worse by less than what it saves, priced by $\lambda$, and a more expensive rival must be better by more than what it costs.
\begin{theorem}[Score certificate]\label{thm:universal}
  Let \( \hat{y} \) be a description with shape \( \hat{s} \) and score \( \hat{q} = Q(\hat{y}) \). Define the floor on the score at each shape
  \[
    \underline{Q}(s) = \underline{E}(s) + \lambda C(s),
  \]
  and set \( q_{\max} = \hat{q} - \min_{s} \underline{Q}(s) \). Then the following hold.
  \begin{enumerate}
    \item Every description \( y' \) with shape \( s' \) satisfies \( Q(y') \geq \underline{Q}(s') \). So a shape \( s' \) can contain a description that beats \( \hat{y} \) only if \( \underline{Q}(s') < \hat{q} \).
    \item The gap to the global minimum satisfies
    \[
      0 \le \hat{q} - \min_{y} Q(y) \leq q_{\max},
    \]
    and \( \hat{y} \) is a global minimiser of $Q$ whenever \( q_{\max} = 0 \).
  \end{enumerate}
\end{theorem}

Theorem~\ref{thm:universal} can be checked for any returned description, regardless of how it was produced, and it is cheap to calculate. The universal floor only needs the singular values and row norms of $W$, so it costs a single SVD. The shapes with \( \underline{Q}(s') < \hat{q} \) are the only places where a better description could still exist, and \( \hat{q} - \underline{Q}(s') \) is the most it could gain there. This is only an upper bound, and it says nothing about whether any description actually reaches it. The certificate also holds only for the given $\lambda$, so a new price means recomputing it. Finally, nothing in the theorem depends on how \( \underline{E} \) was obtained. Any valid lower bound on the error can be swapped in, and a tighter one gives a sharper certificate. Section~\ref{sec:blockwise} builds such a bound.

\begin{figure}
    \centering
    \includegraphics[width=1\linewidth]{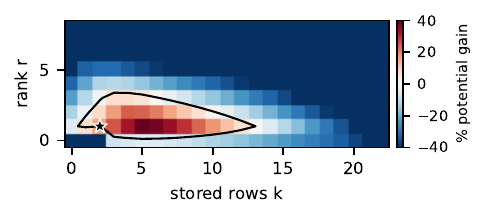}
    \caption{The certificate on a small toy matrix shows where a better description could still exist, and the split floor of Section~\ref{sec:blockwise} closes most of these shapes (Appendix~\ref{sec:howtopart}). {\footnotesize Each shape \( (r, k) \) is coloured by the returned score minus the floor on its score. Red shapes are still open, with the shade giving the most a description there could gain, and blue shapes are pruned. The star is the returned description, and the line bounds the open shapes.}}
    \label{fig:certdemo}
\end{figure}

The certificate prunes much like branch-and-bound (BnB), with each shape playing the role of a box. Each box has a floor on the score of every description inside it, and a box is pruned when its floor is at least the incumbent's score. The box holding the incumbent can never be pruned. Its gap \( \hat{q} - \underline{Q}(\hat{s}) = E(\hat{y}) - \underline{E}(\hat{s}) \) is the slack between the returned error and its own floor, which only a tighter floor can close. Figure~\ref{fig:certdemo} shows the certificate on a small example.

\section{A block-wise tightening of the certificate}\label{sec:blockwise}
Both floors in Corollary~\ref{coro:univfloor} are worst-case bounds, and each can become vacuous for completely benign reasons. The spectral floor forgets which rows are stored, so it fades to zero once \( r + k \) reaches the rank of $W$. The row-energy floor forgets which subspace is used, so it is clipped to zero as soon as a few heavy rows outweigh the entire residual \( T(r) \). Neither failure means the true error is small. Each floor has simply discarded the one piece of information that would have kept it alive. We cannot add that information back without breaking the floors, so we need another route.

The route is to split the rows into blocks. The squared error is a sum over rows, so it is also a sum over blocks, and a floor on each block's share of the error adds up to a floor on the whole. All blocks share the same $L$, but a shared $L$ can fit each block no better than that block's own best rank-$r$ fit, so the universal floor still applies to each block on its own. The split floor therefore reuses the universal floor unchanged. It drops only the requirement that blocks share a basis, keeps the total number of stored rows exact, and so lets each block recover information the universal floor discarded. The only thing we do not know is how the $k$ stored rows are spread over the blocks, so we take the smallest total over every possible spread (Theorem~\ref{thm:splitfloor}).

\begin{figure*}
    \centering
    \includegraphics[width=1\linewidth]{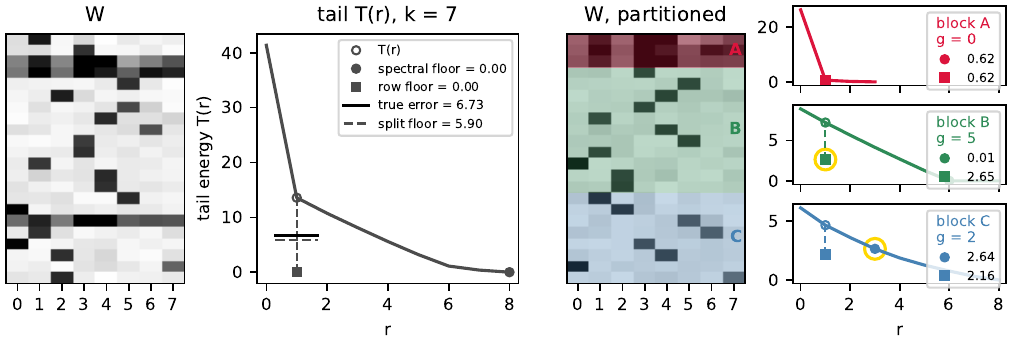}
    \caption{Splitting the rows revives a floor that is exactly zero. Unsplit, the floor is 0. Split, it is \( 0.62 + 2.65 + 2.64 \approx 5.9 \), shown in the left panel just below the true error of 6.73. {\footnotesize A toy matrix at rank 1 with 7 stored rows, unsplit (left) and split into three blocks (right). In each block the binding floor is circled, and no circle means the two floors tie.}}
    \label{fig:blockwise_partition}
\end{figure*}

Figure~\ref{fig:blockwise_partition} shows why this helps on a small matrix at rank 1 with 7 stored rows. Unsplit, both floors are dead. The spectral floor has run out of tail because \( r + k \) reaches the rank of the matrix, and a few heavy rows clip the row-energy floor to zero. Splitting the rows into three blocks changes this. The heavy rows are isolated in one block, which takes none of the stored rows. With them out of the way, the row-energy floor binds in a second block, and the third block only has to store two rows, so its spectral floor binds instead. The total comes back close to the true error.

\begin{theorem}[Split floor]\label{thm:splitfloor}
  Let \( \mathcal{P} = \{ \mathcal{P}_1, \ldots, \mathcal{P}_t \} \) be any partition of the rows of $W$ into $t$ non-empty blocks, and let
  \[
    \mathcal{G}(k) = \Bigl\{ (g_1, \ldots, g_t) : 0 \le g_j \le |\mathcal{P}_j|,\ \textstyle\sum_j g_j = k \Bigr\}
  \]
  be the ways of spreading $k$ stored rows over the blocks. Define the \emph{split floor}
  \[
    \underline{E}_{\mathcal{P}}(r, k) = \min_{(g_1, \ldots, g_t) \in \mathcal{G}(k)} \sum_{j=1}^{t} \underline{E} \bigl( W_{\mathcal{P}_j}; r, g_j \bigr).
  \]
  Then \( E(r, K) \ge \underline{E}_{\mathcal{P}}(r, k) \) for all $r$, $k$ and every $K$ with \( |K| = k \).
\end{theorem}

Every partition gives a valid floor, so we can use several and keep the largest. The partitions may even be chosen by looking at $W$, for example by sorting its rows by norm, because Theorem~\ref{thm:splitfloor} holds for every partition.
\begin{corollary}[Floor over many partitions]\label{coro:splitfloorvalid}
  Let \( \mathcal{P}^{(1)}, \ldots, \mathcal{P}^{(J)} \) be partitions of the rows of $W$, possibly chosen using $W$, and define
  \[
    \underline{E}^{\star}(r, k) = \max \bigl\{ \underline{E}(r, k),\ \underline{E}_{\mathcal{P}^{(1)}}(r, k), \ldots, \underline{E}_{\mathcal{P}^{(J)}}(r, k) \bigr\}.
  \]
  Then \( E(r, K) \ge \underline{E}^{\star}(r, k) \) for every $K$ with \( |K| = k \), and Theorem~\ref{thm:universal} and \eqref{eq:frontierfloor} hold with \( \underline{E}^{\star} \) in place of \( \underline{E} \).
\end{corollary}
Each partition only needs the singular values of its blocks. Together the blocks hold the $n$ rows of $W$, so a partition in total costs at most about one SVD of $W$ to check. Appendix~\ref{sec:howtopart} describes the partitions we use.

There is one catch. The number of ways to spread $k$ rows over $t$ blocks grows like \( \binom{k + t - 1}{t - 1} \), so the minimum in Theorem~\ref{thm:splitfloor} looks expensive. It turns out to be a simple sort. Storing one more row in a block can only lower that block's floor, and by less each time. So the lowest total comes from taking the $k$ largest drops across all blocks, that is the $k$ most negative.
\begin{proposition}[The split floor is a sort]\label{prop:sort}
  Let \( \mathcal{P} \) be a partition as in Theorem~\ref{thm:splitfloor} and \( k \le n \). For each block, write the change in its floor from storing one more row as
  \[
    \delta_j(i) = \underline{E}(W_{\mathcal{P}_j}; r, i) - \underline{E}(W_{\mathcal{P}_j}; r, i - 1), \quad 1 \le i \le |\mathcal{P}_j|.
  \]
  Then \( \underline{E}_{\mathcal{P}}(r, k) = \sum_{j=1}^{t} \underline{E}(W_{\mathcal{P}_j}; r, 0) \) plus the sum of the $k$ smallest, i.e.\ most negative, of all \( \delta_j(i) \).
\end{proposition}

\section{The D-SLR algorithm}\label{sec:algo}
How can every shape be scored without solving each one separately? The key is to fix the basis. If the rank-$r$ fit of the unstored rows is restricted to the top $r$ directions of one fixed basis \( \hat{V} \), then each row's error at rank $r$ is simply its energy outside those directions. The best $k$ rows to store are then the $k$ with the most energy left over, so one sort per rank scores every row count at once. Because the basis is fixed rather than refit, this surrogate error is an upper bound on the true error of the description it proposes.

\textbf{Step 1. Basis and noise.} Compute the SVD of $W$ and flag the rows that fit a low-rank probe of $W$ worst. Compute a second SVD of the unflagged rows only, the same idea as a least trimmed squares fit \cite{rousseeuw1984lts}. Its right singular vectors give the basis \( \hat{V} \), and the bottom of its spectrum gives the noise estimate \( \hat{\tau}^2 \). The flagging only keeps the basis and the noise estimate clean, and has nothing to do with which rows are later stored.

\textbf{Step 2. Sweep.} Score every shape with the surrogate error. The column \( k = 0 \) is instead filled with the exact tail \( T(r) \) from the first SVD, so the truncated SVD is always among the candidates with its true error. This is what makes D-SLR never worse than the truncated SVD at equal cost.

\textbf{Step 3. Pick.} Read a description off the grid, either the lowest error within a parameter limit, the lowest cost that meets an error target, or the lowest score \eqref{eq:score}. By default it uses an EBIC-style price \cite{chen2008ebic}
\begin{equation}\label{eq:ebic}
  \begin{gathered}
    \lambda = \hat{\tau}^2 \bigl( \log(nm) + 2 \log N \bigr), \\
    N = (\min\{n, m\} + 1)(n + 1),
  \end{gathered}
\end{equation}
where $N$ counts every possible shape, so the price does not depend on how much of the grid is computed.

\textbf{Step 4. Final fit.} Store the chosen rows verbatim and fit the rest with one truncated SVD. This gives the exact error \( E(\hat{y}) \) used by the certificate.

The grid and the chosen solution together cost three SVDs regardless of the grid size, and no step iterates. Optionally, the best few shapes can be refit exactly before picking, at one SVD per refit, which corrects any misranking caused by the surrogate's slack near the optimum (Appendix~\ref{sec:algodetails}).

The algorithm has two internal constants, the probe rank \( r_{\mathrm{probe}} \) and the number of flagged rows \( k_{\mathrm{flag}} \), which are fixed across all inputs and appear in neither the problem \eqref{eq:slrobj} nor the certificate (Appendix~\ref{sec:sensitivity}).

\section{Experiments}\label{sec:experiments}
We test D-SLR on synthetic matrices built to stress each part of the method, and on the embedding tables of two LLMs. Costs are counted in SVDs, relative to one SVD of the input matrix. Full settings and further results are in Appendices~\ref{sec:torturedetails}--\ref{sec:realdata}.

\subsection{Synthetic torture test}\label{sec:torturetest}
We compare D-SLR with the methods closest to it in form. Outlier Pursuit is the closest method from the literature, but it targets recovery, so we give it the most generous reconstruction treatment. It stores only the rows it flags, its low-rank part is refit exactly, and its solution is picked with the same score as D-SLR, using D-SLR's own noise estimate. The convex relaxation \eqref{eq:convrelax} targets reconstruction directly. Both are tuned by a grid search over their weights. We also test storing the heaviest rows, and the truncated SVD that D-SLR is meant as a drop-in for. The synthetic test (Appendix~\ref{sec:torturedetails}) has seven cases, one easy and six adversarial, each swept over its generating parameters, for 945 matrices in total.

\begin{table}
\centering
\footnotesize
\setlength{\tabcolsep}{2pt}
\caption{D-SLR found a solution within 1\% of the best score in every run, at a fraction of the cost of the iterative methods. {\footnotesize \textbf{Within 1\%}: share of runs whose score is within 1\% of the best score any method reached on that matrix. \textbf{Worst ratio}: the largest ratio of a method's score to that best score. \textbf{SVDs}: SVDs computed, relative to one SVD of $W$. OP is Outlier Pursuit. \textbf{Time}: median wall-clock time per matrix. The convex relaxation was run on two of the five seeds because of its cost.}}
\label{tab:torture}
\smallskip
\begin{tabular}{@{}lrrrr@{}}
\toprule
Method & Within 1\% & Worst ratio & SVDs & Time (s) \\
\midrule
D-SLR & 100\% & 1.01 & 2.5 & 0.45 \\
OP, tuned & 94\% & 1.03 & 98 & 19.6 \\
OP, default & 81\% & 1.60 & 20 & 4.1 \\
Convex relax. & 78\% & 1.60 & 946 & 248 \\
Heaviest rows & 19\% & 2.86 & 0.8 & 0.2 \\
Truncated SVD & 16\% & 7.66 & 0.3 & 0.06 \\
\bottomrule
\end{tabular}
\end{table}

Table~\ref{tab:torture} shows that D-SLR was within 1\% of the best score in every run, at about 2.5 SVDs and under half a second per matrix. Outlier Pursuit with tuned weights came closest, at 94\% of runs, but needed around 40 times as many SVDs and as much time, and the convex relaxation reached 78\% at over four minutes per matrix.  Storing the heaviest rows is a simplified D-SLR procedure that skips the search and simply orders the rows by size. Without the search it has no surrogate scores, so it must check every row count it considers on the exact error, and on the synthetic matrices it is rarely competitive. The truncated SVD was the cheapest but fell far short whenever some rows fit the shared basis poorly.

On D-SLR's solution, the universal floor alone often left a lot of room, giving about 32\% potential improvement in the median case. The split floor on the other hand, of Section~\ref{sec:blockwise}, certified D-SLR within 1.7\% in the median case (Table~\ref{tab:torturebounds}).

\subsection{LLM embedding tables}
We use D-SLR as a drop-in for the truncated SVD on the embedding tables of Gemma 3 270M \cite{gemmateam2025gemma3} and Llama 3.2 1B \cite{grattafiori2024llama3}. They are among the tallest matrices in an LLM, where rows are cheap compared with ranks (Section~\ref{sec:problem}). Internal weight matrices are closer to square, where the cost argument favours rows far less. This is a showcase rather than a full compression pipeline. Both methods compress the table weighted by calibration statistics from the training split of WikiText-2 \cite{wang2025svdllm, yen2022fwsvd, merity2017wikitext}, so the truncated SVD is close to the compression step of SVD-LLM V2 \cite{wang2025svdllm}. All results are measured on the test split.

\begin{figure}
    \centering
    \includegraphics[width=1\linewidth]{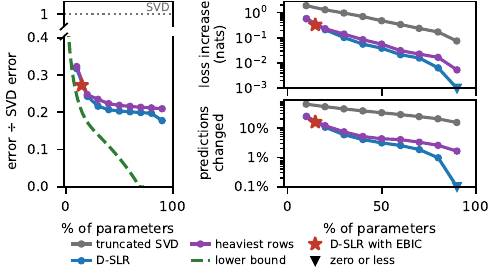}
    \caption{D-SLR preserved the model's behaviour far better than the truncated SVD at every parameter count, on the embedding table of Gemma 3 270M. {\footnotesize \textbf{Left}: Frobenius error relative to the truncated SVD at the same parameter count, with the lower bound from the split floor. \textbf{Top right}: increase in held-out loss. \textbf{Bottom right}: share of held-out predictions that change. The search took two SVDs, and each D-SLR point one more. Heaviest rows stores the rows of largest weighted norm.}}
    \label{fig:gemma3}
\end{figure}

Figure~\ref{fig:gemma3} shows that D-SLR beat the truncated SVD on all three measures at every parameter count, and the same held for Llama 3.2 1B (Appendix~\ref{sec:llmdetails}). At half the table's parameters it changed about ten times fewer predictions, and at 90\% it changed none.

Unlike on the synthetic matrices, storing the heaviest rows is a natural rule here, since the weighting tends to make frequent tokens heavy. We chose its row count from 400 candidates on the exact error. D-SLR used only its own scores, yet matched it at low parameter counts and did slightly better at high ones.

The best achievable error lies between D-SLR and the lower bound. The bound was tightest at high compression where the error is highest and it is needed the most, but loosened gradually. However the gap to it stayed under a quarter of D-SLR's gain over the SVD. On Gemma 3, Outlier Pursuit came close to D-SLR at most limits but needed 174 SVDs and about 70 minutes, against D-SLR's three SVDs and 60 seconds (Appendix~\ref{sec:llmdetails}). On three further real matrices, D-SLR beat the truncated SVD at most parameter counts (Appendix~\ref{sec:realdata}).

\section{Conclusion}\label{sec:conclusion}
We introduced D-SLR, which stores each row verbatim or fits it with a shared low-rank basis, never both. The restriction is free under squared error, makes D-SLR a drop-in for the truncated SVD that is never worse at equal cost, and lets every rank and row count be scored in three SVDs without iteration or tuning, with a certificate that needs no assumptions on the data. It helps most on tall matrices with rows that fit the shared basis poorly, and otherwise returns the truncated SVD. Its default pick assumes the leftover error is noise, and its certificate loosens when many rows and ranks are affordable. In our experiments D-SLR matched the iterative methods at a fraction of their cost and preserved the predictions of LLM embedding tables far better than the truncated SVD. It can also be combined with further compression such as quantisation. Tighter floors and smarter partitions plug straight into the certificate, and better surrogates for the search would tighten the solution itself.

\printbibliography[heading=aistats]

\clearpage
\appendix
\thispagestyle{empty}
\onecolumn
\aistatstitle{D-SLR: The Disjoint Row-Sparse plus Low-Rank Decomposition: \\
Supplementary Materials}

\section{Proof of Proposition~\ref{prop:jointequiv}}
Let \( B = \mathcal{N} \setminus K \). Since $S$ is zero on $B$ and the squared error splits over rows,
\[
  \|W - L - S\|_F^2 = \|W_K - L_K - S_K\|_F^2 + \|W_B - L_B\|_F^2 \ge \|W_B - L_B\|_F^2 \ge T(W_B; r) = E(r, K).
\]
The last inequality is Eckart--Young, which applies because deleting rows cannot raise the rank, so \(
\operatorname{rank}(L_B) \le r \). For the cost, when \( k < n \),
\[
  r(n + m) + km - C(r, k) = rk \ge 0,
\]
with equality only if \( rk = 0 \). When \( k = n \) the convention \( C(r, n) = nm \) makes the difference \( r(n + m) \), which is again positive whenever \( r > 0 \). Finally, every disjoint solution of shape \( (r, k) \) is feasible for \eqref{eq:slrobj}, so the optimum of \eqref{eq:slrobj} is at most \( \min_{|K| = k} E(r, K) \). By \eqref{eq:lossless}, applied with any $K$ of size $k$ that contains the non-zero rows of $S$, it is also at least that, so the two are equal.

\section{Proof of Lemma~\ref{lem:specfloor}}
We prove Lemmas~\ref{lem:specfloor} and~\ref{lem:rowfloor} for any matrix $X$ in place of $W$, since Corollary~\ref{coro:univfloor} and Theorem~\ref{thm:splitfloor} apply them to blocks of rows. Set \( \hat{X} = S + L \), where $S$ stores the rows $K$ exactly and $L$ is the best rank-$r$ fit of the unstored rows $B$, so \( \operatorname{rank}(\hat{X}) \leq \operatorname{rank}(S) + \operatorname{rank}(L) \leq k + r \). By SVD optimality, no matrix of rank at most $r + k$ is closer to $X$ than $T(X; r + k)$,  so \( \|X - \hat{X}\|^2_F \geq T(X; r + k) \). Finally \( \|X - \hat{X} \|^2_F = T(X_B; r) = E(X; r, K)\) because the $K$ rows contribute no error and the residual purely exists in the approximated $B$ rows.
\section{Proof of Lemma~\ref{lem:rowfloor}}
For \( V \subseteq \mathbb{R}^m \) with \( \operatorname{dim}(V) = r \) and some matrix $X$ with rows $x_i$, define:
\begin{equation}\label{eq:rho}
  \rho_i(X; V) = \| x_i \|^2 - \| x_i P_V \|^2
\end{equation}
Two direct consequences that follow are:
\begin{gather}
  0 \leq \rho_i(X; V) \leq \| x_i \|^2 \\
  T(X; r) = \min_{\operatorname{dim}(V) = r} \sum_i \rho_i(X; V) \label{eq:rhotail}
\end{gather}

Using these properties, for every $V$ with $\operatorname{dim}(V) = r$:
\[
  \sum_{i \notin K} \rho_i(V) \;=\; \sum_i \rho_i(V) - \sum_{i \in K} \rho_i(V) \;\geq\; \sum_i \rho_i(V) - \sum_{i \in K} \| x_i \|^2 \;\geq\; \sum_i \rho_i(V) - R(k).
\]

The last step is valid because whichever rows $K$ contains, the energy they carry is at most that of the $k$ heaviest rows of $X$. The subtracted term $R(k)$ is constant in $V$ and $K$ so the inequality is valid after minimising both sides:
\[
  E(r, K) = T(X_B; r) = \min_V \sum_{i \notin K} \rho_i(V) \;\geq\; \min_V \sum_i \rho_i(V) - R(k) = T(r) - R(k).
\]
\section{Proof of Corollary~\ref{coro:univfloor}}
Both Lemmas \ref{lem:specfloor} and \ref{lem:rowfloor} bound the same quantity, $E(r, K)$, depending only on the shape \( |K| = k \), so they bound it uniformly in $K$ and so does the max over them.

Item 2 follows from two facts. The first shows that each mixed floor is itself valid, and the second that the largest of them is an endpoint.

Fix $X$, $r$, and $k$. Then for every $j \in \{0, \ldots, k\}$ and every $K \text{ with } |K| = k$:
\begin{gather}
  E(r, K) \ge T(r + k - j) - R(j), \label{eq:interpless} \\
  \max_{0 \le j \le k} \bigl[ T(r + k - j) - R(j) \bigr] = \max \bigl\{ T(r + k), T(r) - R(k) \bigr\} = \underline{E}(r, k). \label{eq:interpmax}
\end{gather}
The first expression~\eqref{eq:interpless} states that the error is always lower bounded by the interpolating floor. The second expression~\eqref{eq:interpmax} affirms that maximising the interpolated form yields the universal error floor, meaning that the maximum is attained at an endpoint, so no interpolated value exceeds both.

First we prove \eqref{eq:interpless}, the validity of the interpolating floor. Split \( K = K_1 \cup K_2 \) disjointly with \( |K_1| = j,\, |K_2| = k - j \). Using the same notation and steps as in the proof for Lemma~\ref{lem:rowfloor}, for every $r$-dimensional $V$:
\begin{equation}\label{eq:rowfloorsplit}
  \sum_{i \notin K} \rho_i(V) \;=\; \sum_{i \notin K_2} \rho_i(V) - \sum_{i \in K_1} \rho_i(V) \;\geq\; \sum_{i \notin K_2} \rho_i(V) - R(j)
\end{equation}
Minimising \eqref{eq:rowfloorsplit} over $V$ gives \( E(r, K) \;\geq\; T(X_{\mathcal{N} \setminus K_2}; r) - R(j) \), and applying Lemma~\ref{lem:specfloor} to $X$ with stored set $K_2$ gives \( T(X_{\mathcal{N} \setminus K_2}; r) \;\geq\; T(r + k - j) \).

Now we prove \eqref{eq:interpmax}, the maximum over the interpolated floor is either of its endpoints. Write \( T(r + k - j) - R(j) = T(r + k) + f(j) \) where:
\[
  f(j) \;=\; \sum_{i = r+k-j+1}^{r+k} \sigma_i(X)^2 - R(j)
\]
Note that by construction \( f(0) = 0 \) and \( T(r + k) + f(k) = T(r) - R(k) \), that is the endpoints are the floors of Lemmas~\ref{lem:specfloor} and~\ref{lem:rowfloor} respectively. Set $a_1 \geq a_2 \geq \ldots$ as the sorted squared row norms. Identify that \( R(j) = \sum_{i \leq j} a_i \) and:
\[
  f(j) - f(j - 1) = \sigma^2_{r + k - j + 1} - a_j
\]

As $j$ rises, the singular value carries a lower index, hence it is larger in value, while $a_j$ becomes smaller. The increments are therefore non-decreasing, $f$ is convex on \( \{0, \ldots, k\} \), and a convex sequence on consecutive integers is bounded by its endpoints, so \( \max_j f(j) = \max \{0, f(k)\} \).

\section{Proof of Theorem~\ref{thm:universal}}
\begin{enumerate}
    \item Let \( y' \) be a description with shape \( s' \). Its cost is \( C(s') \) and its error is at least \( \underline{E}(s') \) by the universal floor, so \( Q(y') = E(y') + \lambda C(s') \geq \underline{Q}(s') \). If \( \underline{Q}(s') \geq \hat{q} \), then \( Q(y') \geq \hat{q} \), and \( y' \) does not beat \( \hat{y} \). \label{it:cost1}

    \item By item \ref{it:cost1}, every description $y$ with shape $s$ has \( Q(y) \geq \underline{Q}(s) \geq \min_{s'} \underline{Q}(s') \). So \( \min_y Q(y) \geq \min_s \underline{Q}(s) \), that is \( \hat{q} - \min_y Q(y) \leq q_{\max} \). The gap is at least $0$ because \( \hat{y} \) is itself a description. If \( q_{\max} = 0 \), the gap is therefore $0$, and \( \hat{y} \) is a global minimiser.
\end{enumerate}
The bounds also cover overlapping solutions of \eqref{eq:slrobj}, priced at \( r(n + m) + km \) parameters. By Proposition~\ref{prop:jointequiv}, such a solution has error at least \( E(r, K) \) and costs at least \( C(r, k) \), so its score is at least \( Q(r, K) \), the score of a description.

\section{Proof of Theorem~\ref{thm:splitfloor}}
Use the same definition of $\rho_i(V)$ as in \eqref{eq:rho}. Set the allocation \( (g_1, \ldots, g_t) \text{ s.t. } g_j = |\mathcal{P}_j \cap K|\). For any $r$-dimensional $V$, the sum over the residual energy $\rho_i(V)$ for rows in $B$ can be expressed as the double sum over each row in every block $\mathcal{P}_j$:

\begin{equation}\label{eq:blockrho}
  \sum_{i \in B} \rho_i(V) = \sum_{j=1}^t \sum_{i \in \mathcal{P}_j \setminus K} \rho_i(V)
\end{equation}

If one decouples each block's subspace from the shared subspace $V$ and minimises over each one individually, the sum at the shared subspace must be at least as large:

\[
  \begin{aligned}
    \sum_{j=1}^t \sum_{i \in \mathcal{P}_j \setminus K} \rho_i(V) &\geq \sum_{j=1}^t \min_{\operatorname{dim}(V') = r} \sum_{i \in \mathcal{P}_j \setminus K} \rho_i(V') = \sum_{j=1}^t T(W_{\mathcal{P}_j \setminus K}; r) \\
    &= \sum_{j=1}^t E(W_{\mathcal{P}_j}; r, K \cap \mathcal{P}_j) \geq \sum_{j=1}^t \underline{E}(W_{\mathcal{P}_j}; r, g_j)
  \end{aligned}
\]

The first equality is borrowed from \eqref{eq:rhotail}, and the last inequality is just the universal floor of Corollary~\ref{coro:univfloor} applied to each block in the sum individually.

Now the right-hand side no longer depends on the shared subspace $V$, so we can minimise the left-hand side over all subspaces and still retain the inequality:
\[
  E(r, K) = \min_{\operatorname{dim}(V)=r} \sum_{i \in B} \rho_i(V) \geq \sum_{j=1}^t \underline{E}(W_{\mathcal{P}_j}; r, g_j)
\]

Finally because \( (g_1, \ldots, g_t) \) is an element of $\mathcal{G}(k)$, minimising over all allocations can only lower it:
\[
  E(r, K) \geq \min_{(g_1', \ldots, g_t') \in \mathcal{G}(k)} \sum_{j=1}^t \underline{E}(W_{\mathcal{P}_j}; r, g_j') = \underline{E}_{\mathcal{P}}(r, k)
\]
\section{Proof of Corollary~\ref{coro:splitfloorvalid}}
Each term in the set that is being maximised over is a valid lower bound on $E(r, K)$. The first by Corollary~\ref{coro:univfloor} and the rest by Theorem~\ref{thm:splitfloor}. The latter places no condition on how each partition is created. A maximum of finitely many valid lower bounds is also a valid lower bound, and Theorem~\ref{thm:universal} accepts any valid floor.
\section{Proof of Proposition~\ref{prop:sort}}
First we establish that each block's floor is non-increasing and convex in $k$. Write the floors of Lemmas~\ref{lem:specfloor} and~\ref{lem:rowfloor} as functions of $k$, \( e_1(X; k) = T(X; r + k) \) and \( e_2(X; k) = T(X; r) - R(X; k) \) respectively. Their respective increments are \( \Delta_1(X; k) = -\sigma_{r + k + 1}(X)^2 \) and \( \Delta_2(X; k) = -a_{k + 1} \) where \( a_1 \geq a_2 \geq \ldots \) are the sorted squared row norms. Both $\Delta_1 \leq 0$ and $\Delta_2 \leq 0$ for any $X$ and $k$, so as a sequence in $k$ they both are non-increasing, while their respective sequences of increments are non-decreasing as the singular values and squared row norms are sorted. Therefore $e_1$ and $e_2$ are non-increasing and convex. Furthermore a pointwise max of non-increasing, convex sequences, together with the point 0, is itself non-increasing and convex.

\paragraph{$\geq$} Any allocation \( (g_1, \ldots, g_t) \in \mathcal{G}(k) \) gives the total \( \sum_j \underline{E}(W_{\mathcal{P}_j}; r, 0) + \sum_j \sum_{i \le g_j} \delta_j(i) \). The double sum picks $k$ distinct increments, so it is at least the sum of the $k$ smallest. This direction does not use convexity, so the sorted value is a valid floor for any blocks.

\paragraph{$\leq$} Take the $k$ smallest increments, breaking ties in favour of the smaller $i$, then the smaller $j$. If $\delta_j(i)$ is picked but $\delta_j(i-1)$ is not, then \( \delta_j(i-1) < \delta_j(i) \) is impossible because the smaller increment would have been picked first, \( \delta_j(i-1) > \delta_j(i) \) is impossible by convexity, and equality is excluded by the tie-breaking rule. So the increments picked from block $j$ are its first $g_j$, with \( g_j \le |\mathcal{P}_j| \) because the block has no further increments. This is an allocation in \( \mathcal{G}(k) \), and its total is exactly the right-hand side.

\section{How to best partition the matrix}\label{sec:howtopart}
Theorem~\ref{thm:splitfloor} holds for any partition, so which partitions give a tight floor? Each of the two floors can be helped in its own way.

\paragraph{Spectral floor.} Split the rows into $t$ blocks that each look like a smaller copy of $W$, so each block holds about a $1/t$ share of its spectrum. If the $k$ stored rows are spread evenly, each block concedes only $k/t$ extra directions instead of $k$, and
\[
  \sum_{j} T\bigl(W_{\mathcal{P}_j}; r + k/t\bigr) \approx T(W; r + k/t) \ge T(W; r + k).
\]
The split floor takes the worst spread, not the even one. But each block's floor is convex in its number of stored rows (proof of Proposition~\ref{prop:sort}), so when the blocks are alike the even spread is the worst. A block also needs enough rows. With few rows it has few singular values, and its floor reaches zero after only a few ranks or stored rows.

\paragraph{Row-energy floor.} The row floor \( T(r) - R(k) \) is zero as soon as the $k$ heaviest rows hold more energy than the tail. Putting the heaviest rows in a block of their own confines this. That block's floor drops to zero, but the other blocks only subtract their own lighter rows.

\begin{figure}
    \centering
    \includegraphics[width=1\linewidth]{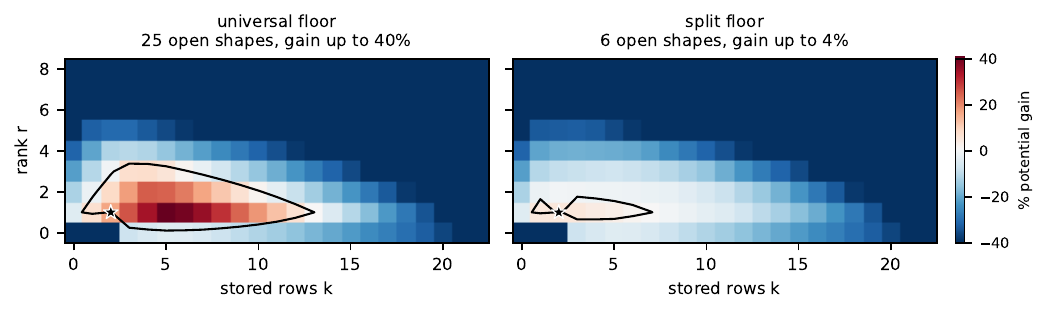}
    \caption{The certificate on a small toy matrix. It shows where a better description could still exist, and how much a tighter floor narrows this down. The left panel uses the universal floor and the right panel the split floor of Section~\ref{sec:blockwise}, which closes most of the open shapes. {\footnotesize Each shape \( (r, k) \) is coloured by the returned score minus the floor on its score. Red shapes are still open, with the shade giving the most a description there could gain, and blue shapes are pruned. The star is the returned description, and the black line separates open from pruned shapes.}}
    \label{fig:certdemosplit}
\end{figure}

\paragraph{What we use.} Sorting the rows by norm and dealing them round-robin into blocks gives each block a similar mix of heavy and light rows. We do this starting with \( t = \lfloor n / (0.8 m) \rfloor \) blocks and halving $t$ up to four times. A hybrid variant serves both floors. The sorted rows are cut into two or three strata of growing size, and each stratum is dealt round-robin into blocks of about $0.8m$ or $1.5m$ rows. The heaviest rows then share a small block, and the blocks within each stratum stay alike. All experiments use these partitions together with the unsplit floor. These are heuristics, but a poor partition costs nothing, because Corollary~\ref{coro:splitfloorvalid} always keeps the unsplit floor. Figure~\ref{fig:certdemosplit} shows the effect on a small matrix.

\section{Algorithm details}\label{sec:algodetails}
\paragraph{Defaults.} The probe rank is \( r_{\mathrm{probe}} = \min\{15, \min\{n, m\} - 1\} \), and half of the rows are flagged, \( k_{\mathrm{flag}} = \lfloor n/2 \rfloor \). Flagging is deliberately over-inclusive, because a flagged row is only left out of the basis fit and can still be stored or approximated like any other.

\paragraph{Noise estimate.} \( \hat{\tau}^2 \) is the mean of the bottom half of the squared singular values of the unflagged block, divided by its larger dimension. On pure noise the bottom of the spectrum sits below the average, so this raw estimate is low by a factor that depends only on the block's aspect ratio. We compute that factor once from the Marchenko--Pastur law and divide it out.

\paragraph{Price.} The price \eqref{eq:ebic} is adapted from EBIC \cite{chen2008ebic} rather than taken from it. Like BIC, it charges \( \log(nm) \) per parameter, for the $nm$ observed entries. EBIC adds twice the log of the number of candidate models of the same size, once per model. Bounding that number by \( p^{\nu} \) for $\nu$ parameters drawn from $p$ candidates turns it into \( 2 \log p \) per parameter, and we use the number of shapes $N$ for $p$. Both terms are multiplied by \( \hat{\tau}^2 \) to put them in squared-error units. This is a heuristic choice, and any other price can be used, because the certificate holds for every $\lambda$.

\paragraph{Shortlist verification.} With verification enabled, the shortlist by default holds the eight best-scoring shapes with \( k > 0 \), the best shape with \( k = 0 \), plus the best row count for each of four ranks on a geometric ladder. Each shape with \( k > 0 \) is refit exactly twice, once on the rows the sweep proposes and once on its heaviest rows, at one SVD of the unstored block each, and the better of the two is kept. The shortlist is then re-ranked on the exact score. Shapes with \( k = 0 \) are already exact and cost nothing.

\paragraph{Other candidates.} Storing the heaviest rows is a simplified D-SLR procedure that skips the search and simply orders the rows by size, so the certificate applies to its solutions unchanged. Without the search it has no surrogate scores, so on its own it must check every row count it considers on the exact error, and in our synthetic tests it is rarely competitive (Table~\ref{tab:torture}).

\section{Synthetic torture test}\label{sec:torturedetails}
All timings in this paper come from one consumer workstation with an AMD Threadripper 2950X processor and 128\,GB of memory, with the same BLAS thread settings for every method and experiment. A GPU computed only the calibration statistics and held-out measures of the LLMs, which no timing includes.

Two torture tests are run, one that compares all methods, and one that only compares the faster methods. The first test is more varied overall but was restricted in certain aspects to save time. The second test explores what was left out.

For the main test used in Section~\ref{sec:torturetest} the generated matrix follows a common recipe that is then changed. It is a rank-10 background plus some heavy rows and noise, at \( 4000 \times 200 \). The background is a product of Gaussian factors with entries of unit variance, the heavy rows are Gaussian rows scaled up and placed at random, and the noise is Gaussian. The core grid varies the share of heavy rows (2, 10 and 20\%), the noise level per entry (0.05, 0.1 and 0.2) and how much heavier the heavy rows are (1.5, 4 and 16 times), for 27 settings. Each case changes one thing about this recipe.
\begin{itemize}
  \item \textbf{Plain}: the recipe as is.
  \item \textbf{Many heavy rows}: 20, 30 and 40\% of the rows are heavy.
  \item \textbf{Mixed heavy sizes}: each heavy row gets its own size, between 0.25 and 32 times the base size.
  \item \textbf{Aligned heavy rows}: the heavy rows all point in one direction.
  \item \textbf{Uneven background}: each background row is scaled by a random weight from a Pareto distribution with shape 1.5, so the background rows have very uneven norms.
  \item \textbf{Heavy-tailed noise}: the noise follows a Student $t$ distribution with 3 degrees of freedom.
  \item \textbf{Partly heavy rows}: only 30\% of each heavy row's entries are large.
\end{itemize}
With five seeds this gives 945 matrices. Every method returns a description of the form \( \hat{W} = S + L \) and picks it with the EBIC score \eqref{eq:ebic} at D-SLR's noise estimate, since none of the baselines has a reliable estimate of its own. Each pick is then scored with the true noise level, which is known here. Outlier Pursuit flags the rows whose sparse part exceeds \( 3 \hat{\tau} \sqrt{m} \), with \( \hat{\tau}^2 \) D-SLR's noise estimate. For each weight, the flagged rows are stored and the rest are refit exactly at every rank. Its default weight is the universal weight of PCP \cite{candes2011robust}, rescaled by \( \sqrt{m} \) to act on row norms, giving \( \sqrt{m} / \sqrt{\max\{n, m\}} \). The convex relaxation is solved by ADMM, refit and picked in the same way, and because of its cost was run on two of the five seeds. It did worse than Outlier Pursuit, likely because it must select both the rows and the rank through its penalties, while Outlier Pursuit has its rank picked by the same score as D-SLR. Storing the heaviest rows is refit exactly at every rank, on a grid of 48 row counts. The time of Outlier Pursuit at its default weight is estimated from that weight's share of the iterations in the tuned run. The best score on a matrix is the lowest score of any method run on it.

The second test aims to target parameters that could affect D-SLR more than others. A smaller version of the first test was run but now varying the matrix shape and background rank. Table~\ref{tab:tortureshapes} repeats the comparison at three shapes (\( 2000 \times 200 \), \( 4000 \times 200 \) and \( 4000 \times 400 \)) and eight background ranks (1, 4, 8, 12, 16, 32, 64 and 96), at each case's middle share of heavy rows and five seeds, for 840 matrices. Only the methods cheap enough to run on every matrix are included. D-SLR's results held in similar fashion to the first test.

\begin{table}
\centering
\footnotesize
\setlength{\tabcolsep}{2pt}
\caption{Every method on the torture test, including two variants of D-SLR. Without flagging, the heavy rows pull the basis away from the background, which is why Step~1 flags them. {\footnotesize Columns as in Table~\ref{tab:torture}. \textbf{Verified} adds the shortlist refits of Appendix~\ref{sec:algodetails}. \textbf{Default} uses the method's default weight instead of a grid search.}}
\label{tab:torturefull}
\smallskip
\begin{tabular}{@{}lrrr@{}}
\toprule
Method & Within 1\% & Worst ratio & SVDs \\
\midrule
D-SLR & 100\% & 1.01 & 2.5 \\
D-SLR, verified & 100\% & 1.00 & 8.2 \\
D-SLR without flagging & 61\% & 5.96 & 2.9 \\
Outlier Pursuit, tuned & 94\% & 1.03 & 98 \\
Outlier Pursuit, default & 81\% & 1.60 & 20 \\
Convex relaxation, tuned & 78\% & 1.60 & 946 \\
Convex relaxation, default & 12\% & 12.04 & 21 \\
Heaviest rows & 19\% & 2.86 & 0.8 \\
Truncated SVD & 16\% & 7.66 & 0.3 \\
\bottomrule
\end{tabular}
\end{table}

\begin{table}
\centering
\footnotesize
\setlength{\tabcolsep}{3pt}
\caption{D-SLR stayed within 1.1\% of the best score at every shape and background rank, and within 1\% in all but one of the 840 runs. {\footnotesize Table~\ref{tab:torturefull} varies the share, size and noise of the heavy rows at one shape and background rank. This table holds those at each case's middle setting and varies the shape and background rank instead. Only the methods cheap enough to run on every matrix are included, so the best is taken over the methods shown. Columns as in Table~\ref{tab:torture}.}}
\label{tab:tortureshapes}
\smallskip
\begin{tabular}{@{}lrrr@{}}
\toprule
Method & Within 1\% & Worst ratio & SVDs \\
\midrule
D-SLR & 99.9\% & 1.01 & 2.5 \\
D-SLR, verified & 100\% & 1.01 & 7.9 \\
D-SLR without flagging & 63\% & 5.57 & 2.9 \\
Heaviest rows & 25\% & 1.78 & 1.1 \\
Truncated SVD & 14\% & 8.29 & 0.3 \\
\bottomrule
\end{tabular}
\end{table}

\begin{table}
\centering
\footnotesize
\caption{The split floor certified D-SLR's solution far more tightly than the universal floor on the torture test. {\footnotesize Each value is the most any solution could improve on the score of D-SLR's pick, at the price from its own noise estimate, by Theorem~\ref{thm:universal} with the given floor, as a percentage of that score. Over the 945 matrices of Table~\ref{tab:torturefull}.}}
\label{tab:torturebounds}
\smallskip
\begin{tabular}{lrr}
\toprule
Bound & Median (\%) & Worst (\%) \\
\midrule
Universal & 32.0 & 79.1 \\
Split & 1.7 & 20.5 \\
\bottomrule
\end{tabular}
\end{table}

\section{LLM embedding tables}\label{sec:llmdetails}
The embedding tables are \( 262144 \times 640 \) for Gemma 3 270M and \( 128256 \times 2048 \) for Llama 3.2 1B, about 63\% and 21\% of each model's parameters. Both models share this table between the input embedding and the output head. The tables are weighted using the forward activations and the curvature of the output loss, as in \cite{wang2025svdllm, yen2022fwsvd}, so that the reconstruction error better approximates the downstream error (Appendix~\ref{sec:weighted}). Calibration uses about 33k tokens of the train split of WikiText-2 for Gemma 3 and 130k for Llama 3.2, and evaluation uses about 16k tokens of its test split. Both methods see the same weighted table, and each parameter limit is a share of the table's parameters.

The held-out loss is the mean cross-entropy per token in nats, and its increase is measured against the uncompressed model. A prediction changes when the model's most likely next token differs from that of the uncompressed model.

\begin{figure}
    \centering
    \includegraphics{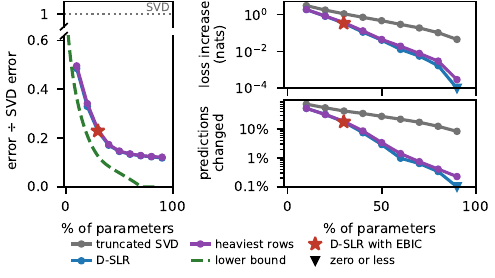}
    \caption{The embedding table of Llama 3.2 1B, as in Figure~\ref{fig:gemma3}. D-SLR again beats the truncated SVD on all three measures at every parameter count.}
    \label{fig:llama32}
\end{figure}

\begin{figure}
    \centering
    \includegraphics{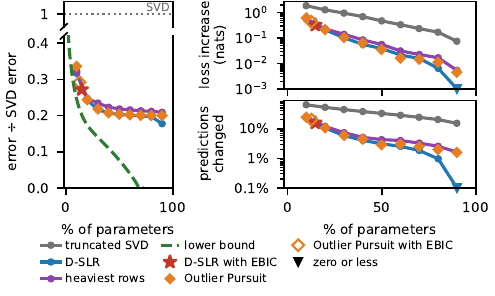}
    \caption{Figure~\ref{fig:gemma3} with Outlier Pursuit added. It comes close to D-SLR at most limits but needs 174 SVDs against D-SLR's three. Its weight only steers the number of stored rows indirectly, so at the highest limits it can only add rank, and it falls behind.}
    \label{fig:gemma3op}
\end{figure}

The D-SLR points are the standard algorithm of Section~\ref{sec:algo}, without shortlist verification. Figure~\ref{fig:llama32} shows Llama 3.2 1B, and Figure~\ref{fig:gemma3op} adds Outlier Pursuit to Gemma 3. Outlier Pursuit was given D-SLR's noise estimate and run at six weights plus the empty row set, with at most 50 iterations each, where the torture test never needed more than 26. Its description at each limit is the lowest exact error among all its candidates within that limit.

\paragraph{Assets.} Gemma 3 is released under the Gemma Terms of Use and Llama 3.2 under the Llama 3.2 Community License. WikiText-2 is released under CC BY-SA \cite{merity2017wikitext}.

\section{Further real-world matrices}\label{sec:realdata}
We also used D-SLR as a drop-in for the truncated SVD on three real matrices with no model in the loop, so the error is plain squared error. PaviaU is a hyperspectral image of Pavia University \cite{paviau}, with one row per pixel and one column per spectral band, giving a \( 207400 \times 103 \) matrix that we centre per band. GEANT holds the traffic of a European research network \cite{uhlig2006geant}, with one row per 15-minute snapshot and one column per origin-destination flow, giving a \( 10769 \times 529 \) matrix that we use as is. The electricity data holds the load of Portuguese customers every 15 minutes \cite{trindade2015electricity}. We keep 2012 to 2014 and the 321 customers active from the start, centre and scale each customer, and cut the series into daily load curves, the usual unit in load-profile analysis. This gives a \( 351816 \times 96 \) matrix with one row per customer and day. A stored row is a pixel, snapshot or customer-day kept exactly, and neither method is charged for the means removed by centring. The lower bound uses the partitions of Appendix~\ref{sec:howtopart}.

Figure~\ref{fig:scenes} shows that D-SLR matched or beat the truncated SVD at every limit on all three matrices. On PaviaU it gained from 3\% of the parameters upward. PaviaU and GEANT are close to low rank, so each added rank removes much of the remaining error and it is difficult for a few stored rows to do better, which keeps their gains modest. On GEANT D-SLR gained at the tightest limits, where a few snapshots fit the shared basis poorly and a stored row costs about a twentieth of a rank. From 5\% to 80\% it returned the truncated SVD, as did the EBIC pick, and at 90\% it stored rows again. The lower bound shows that room remained there, and the shortlist refits of Appendix~\ref{sec:algodetails} found it, at about 27 SVDs per limit instead of one. On the electricity data D-SLR beat the truncated SVD at every limit, by storing individual customer-days that fit the typical daily shapes poorly, and the gain grew with the parameter count. Put another way, to reach the truncated SVD's error D-SLR needed up to a quarter fewer parameters on PaviaU and on the electricity data, and up to half on GEANT at its tightest limits. On all three matrices the lower bound fades at the most generous limits, where many rows and ranks are affordable and the floors loosen.

\begin{figure*}
    \centering
    \includegraphics[width=\textwidth]{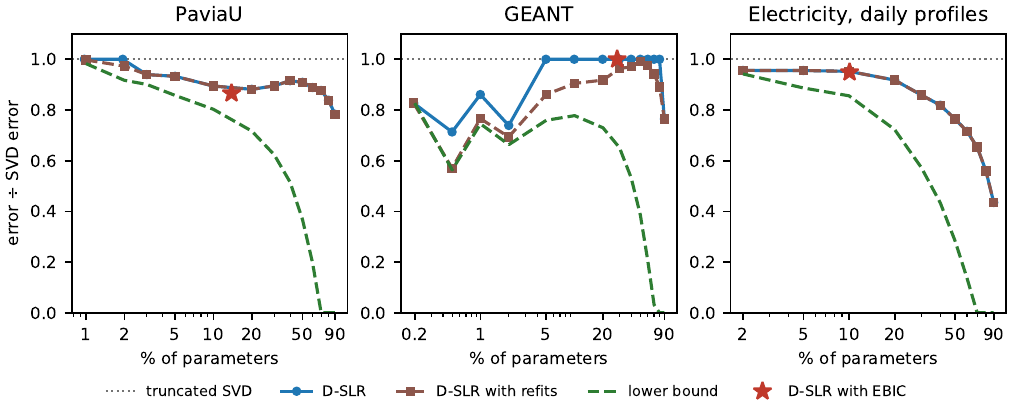}
    \caption{D-SLR matched or beat the truncated SVD at every parameter count on all three matrices. {\footnotesize Error relative to the truncated SVD at the same parameter count, with the lower bound from the split floor. The dashed line adds the shortlist refits, and covers the solid line where the two agree. The star is the EBIC pick, which on GEANT stores no rows and so equals the truncated SVD.}}
    \label{fig:scenes}
\end{figure*}

\paragraph{Assets.} The electricity data is released under CC BY 4.0 \cite{trindade2015electricity}. The PaviaU scene \cite{paviau} and the GEANT traffic matrices \cite{uhlig2006geant} are distributed by their authors for research use, with no stated license. No data is redistributed with this paper.

\section{Sensitivity to $r_{\mathrm{probe}}$ and $k_{\mathrm{flag}}$}\label{sec:sensitivity}
Unlike the weights in \eqref{eq:convrelax}, the constants \( r_{\mathrm{probe}} \) and \( k_{\mathrm{flag}} \) appear in neither the problem \eqref{eq:slrobj} nor the certificate, which holds whatever price is used. They only affect which solution is proposed and the noise estimate \( \hat{\tau}^2 \). And unlike a penalty weight, \( \tau^2 \) has a true value in the data that the estimate can be checked against.

We ran D-SLR at every pair of \( r_{\mathrm{probe}} \in \{0, 1, 2, 4, 8, 15, 30, 60, 100\} \) and \( k_{\mathrm{flag}} / n \in \{0, 0.02, 0.05, 0.1, 0.2, 0.3, 0.5, 0.7, 0.9\} \), on the seven torture-test cases at three settings each. These are the middle of the core grid, its hardest corner, with heavy rows only 1.5 times the background at the highest noise, and a background of rank 32, more than the default probe rank. With three seeds this gives 63 matrices of size \( 4000 \times 200 \). Each result is scored with the true noise level and compared with the defaults on the same matrix.

\begin{figure*}
    \centering
    \includegraphics[width=\textwidth]{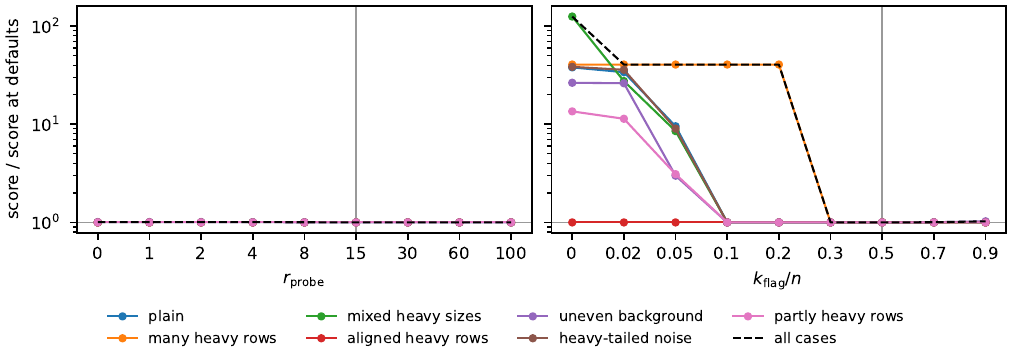}
    \caption{The probe rank barely mattered, and the flagged share only failed when no more rows were flagged than the matrix had heavy rows. {\footnotesize Worst score over each case's matrices relative to the score at the defaults (grey lines), varying one constant at a time. \textbf{Left}: \( r_{\mathrm{probe}} \) at \( k_{\mathrm{flag}} = n/2 \). \textbf{Right}: \( k_{\mathrm{flag}} \) at \( r_{\mathrm{probe}} = 15 \).}}
    \label{fig:sensitivity}
\end{figure*}

Figure~\ref{fig:sensitivity} shows that the probe rank barely mattered. From 0 to 100 no matrix scored more than 1\% worse than at the defaults, including those whose background has more directions than the probe. The flagged share mattered only when it was too small. Every setting more than 5\% worse than the defaults flagged no more rows than the matrix had heavy rows. The extra heavy rows then entered the basis fit, the noise estimate and with it the price came out many times too high, and D-SLR stored too little. At the default probe rank, from 30\% of the rows upward no matrix scored more than 3\% worse. The default flags half the rows, so it can only fail when more than half the rows fit the shared basis poorly, and then the matrix is no longer a low-rank background with some rows apart. On every matrix the defaults were also within 1\% of the best setting on the grid.

\section{Weighted inputs}\label{sec:weighted}
D-SLR minimises squared error, but some uses call for a weighted error. We support weights that scale each row and a Gram matrix that mixes the columns, and minimise
\[
  \sum_{i} w_i \, (x_i - \hat{x}_i) \, G \, (x_i - \hat{x}_i)^\top
\]
over the rows $x_i$ of $W$, with \( w_i > 0 \) and $G$ positive semidefinite. We add a small ridge \( \epsilon I \) to $G$ and floor the weights at \( 10^{-6} \) of their largest value, so that both maps below can be inverted, and minimise the objective with \( G + \epsilon I \) in place of $G$. Writing \( G + \epsilon I = H H^\top \) with $H$ its square Cholesky factor, this is the plain squared error of \( M = \operatorname{diag}(\sqrt{w}) \, W H \), so D-SLR runs unchanged on $M$ and its solution is mapped back. Both maps act within a row or on the right, never across rows, so a stored row of $M$ is a stored row of $W$ and still costs $m$ parameters. A full covariance across rows would mix them, and a stored row would no longer be a row of $W$. The stored rows are copied from $W$ itself, so they stay exact, and the low-rank part is mapped back by undoing the row scale and $H$. The floors and the certificate hold unchanged on $M$, so they bound this weighted error, which exceeds the error under $G$ alone by \( \epsilon \sum_i w_i \|x_i - \hat{x}_i\|^2 \).

For the LLM tables, $G$ is the mean outer product of the hidden states entering the output head, and \( w_i \) the mean of \( p_i (1 - p_i) \), where \( p_i \) is the predicted probability of token $i$. With these values, the weighted sum is just the second-order approximation of the cross-entropy loss, when dropping the first-order term. This keeps only the diagonal of the curvature across tokens and treats it as independent of the hidden states, so it is an approximation. The embedding table is shared between input and output and the decomposed table is used to replace both, and the held-out loss measures the result.

\end{document}